\pdfoutput=1
\documentclass[10pt, a4paper]{article}

\usepackage{lrec-coling2024} 

\usepackage{booktabs}
\usepackage{multirow}

\title{Transferring BERT Capabilities from High-Resource to Low-Resource Languages Using Vocabulary Matching}

\name{
Piotr Rybak
} 

\address{
Institute of Computer Science, Polish Academy of Sciences\\
ul. Jana Kazimierza 5, 01-248 Warsaw, Poland\\
\texttt{piotr.rybak@ipipan.waw.pl}
}

\abstract{
Pre-trained language models have revolutionized the natural language understanding landscape, most notably BERT (Bidirectional Encoder Representations from Transformers). However, a significant challenge remains for low-resource languages, where limited data hinders the effective training of such models. This work presents a novel approach to bridge this gap by transferring BERT capabilities from high-resource to low-resource languages using vocabulary matching. We conduct experiments on the Silesian and Kashubian languages and demonstrate the effectiveness of our approach to improve the performance of BERT models even when the target language has minimal training data. Our results highlight the potential of the proposed technique to effectively train BERT models for low-resource languages, thus democratizing access to advanced language understanding models.
 \\ \newline  }

\begin{document}

\maketitleabstract

\section{Introduction}
\label{sec:intro}
The field of natural language processing (NLP) has made remarkable progress in recent years, largely due to the rise of pre-trained language models. These models, most notably BERT (Bidirectional Encoder Representations from Transformers, \citealp{devlin-etal-2019-bert}), leverage unsupervised pre-training on massive text corpora to learn meaningful contextual representations of words. As a result of pre-training, such models have significantly reduced the data and computational requirements for further fine-tuning to a specific downstream task and have demonstrated exceptional capabilities in a wide range of NLP tasks, from text classification, named entity recognition, question answering to information retrieval.

While pre-trained language models have ushered in a new era of NLP, a significant challenge remains in extending their benefits to low-resource languages. The limited availability of text data and linguistic resources makes it difficult or impossible to pre-train models such as BERT using traditional methods. Moreover, these languages are often overlooked by the NLP community, leaving their speakers at a disadvantage in the era of AI-driven technologies.

In response to this challenge, our work proposes a simple approach of transferring BERT capabilities from high-resource and low-resource languages. The core idea is to use an external dictionary to adapt the BERT tokenizer of the high-resource model to properly initialize the BERT model for the low-resource language. This process makes further fine-tuning much more efficient and allows the BERT model to be trained with minimal training data. 

We demonstrate the effectiveness of our approach by conducting experiments on two low-resource languages: Silesian and Kashubian. These languages, spoken by relatively small communities, face significant data scarcity and have been underrepresented in the NLP research community. Silesian is a West Slavic ethnolect spoken primarily in the region of Upper Silesia in Poland and is considered either one of the four major dialects of Polish or a separate regional language distinct from Polish. Kashubian is also a West Slavic language but is spoken primarily in the Pomeranian Voivodeship of Poland. Similar to Silesian, Kashubian is recognized either as a Polish dialect or as a separate language.

\vspace{4pt}
To summarize, our contributions are: 
\begin{enumerate}
    \item Release of the first BERT models for Silesian and Kashubian,\footnote{The models are available at: \url{https://hf.co/ipipan/kashubian-herbert-base} and \url{https://hf.co/ipipan/silesian-herbert-base}}
    \item A simple yet effective method to transfer BERT capabilities from high-resource to low-resource languages.
\end{enumerate}

\section{Related Work}
\label{sec:related}

\subsection{Training BERT models for low-resource languages}
The development of multilingual models, including BERT, has been an important step in addressing language diversity. However, training multilingual models does not necessarily ensure good performance for low-resource languages \citep{wu-dredze-2020-languages,lauscher-etal-2020-zero} or robust alignment of word embeddings across languages \citep{Cao2020Multilingual}.

A significant line of research has focused on training cross-lingual models, using both supervised and unsupervised methods. Supervised approaches use parallel data to train models, either by simply pre-training on parallel data \citep{lample2019cross} or by aligning the output representation of the parallel sentences \citep{feng-etal-2022-language}. Unsupervised approaches often mine word or sentence pairs to be used later in the training phase \citep{schuster-etal-2019-cross,hangya-etal-2022-improving}.

Techniques to improve performance specifically for low-resource languages include increasing the BERT vocabulary \citep{wang-etal-2020-extending}, reducing the number of parameters, and pre-training with multiple targets \citep{gessler-zeldes-2022-microbert}. Other approaches focus on adapting the BERT model embeddings either using parallel data \citep{tran2020english} or simply by matching tokens between tokenizers' vocabularies \citep{arkhipov-etal-2019-slavic}. Others use dictionaries to weakly translate datasets used for pre-training \citep{wang-etal-2022-expanding}.

\subsection{Silesian and Kashubian resources}
Silesian and Kashubian, our focus languages, have very limited linguistic resources and training data. The only publicly available corpus is Wikipedia.

WikiANN \citep{rahimi-etal-2019-massively}, a multilingual named entity recognition dataset automatically generated from Wikipedia articles, contains 300 examples for both Silesian and Kashubian. The manual inspection shows its low quality, as most of the examples are just article titles.

Tatoeba\footnote{\url{https://tatoeba.org/}} is a crowd-sourced collection of user-provided translations in a large number of languages, including 61 sentences for Silesian and 962 sentences for Kashubian. They are useful for evaluating machine translation models, but not language models.

\section{Method}
\label{sec:method}

\subsection{Data collection and cleaning}
We use the \texttt{20230901} snapshot of Silesian and Kashubian Wikipedia as a corpus for pre-training the BERT model. To improve the data quality, we perform several steps to clean the raw text. First, we parse each article using WikiExtractor \citep{Wikiextractor2015}, but contrary to the default settings, we keep lists as valid text. We remove short and repetitive articles about villages, cities, provinces, etc., as well as empty articles about a particular day or year. For Silesian, we also remove over 40,000 automatically generated articles about plants and convert articles written in \emph{Steuer} script to \emph{Ślabikorz} script using the Silling converter.\footnote{\url{https://silling.org/konwerter-ze-steuera-na-slabikorz/}} The final corpus consists of 485,736 words for Silesian and 300,497 words for Kashubian (see Table \ref{tab:data}).

\begin{table}[!ht]
\renewcommand*{\arraystretch}{1.3}
\setlength{\tabcolsep}{8pt}
\centering
\begin{tabular}{l|rrr}
    \toprule
    \bf{Language} & \bf{\# Docs} & \bf{\# Words} & \bf{\# Dict} \\
    \midrule
    Silesian & 5,068 & 485,736 & 23,934 \\
    Kashubian & 3,061 & 300,497 & 23,762 \\
    \bottomrule
\end{tabular}
\caption{Number of documents and words in the training corpus, and the size of the bilingual dictionary used for vocabulary matching.} 
\label{tab:data}
\end{table}

\subsection{Vocabulary matching}
\label{sec:vocab}
Pre-trained BERT models serve as a perfect initialization for training models for new languages, especially when the languages are closely related. While the weights for the self-attention layers can be copied directly from the source to the target model, the challenge arises for the embedding layer and the final masked language modeling (MLM) classification head, as both are tightly coupled with the tokenizer.

To overcome the difference in the tokenizers' vocabularies, we used a method similar to \citep{arkhipov-etal-2019-slavic}, but extended it to the cross-lingual scenario. If a token from the target model vocabulary is present in the bilingual dictionary and its translation is present in the source model vocabulary, we directly copy its weights. For example, the token \texttt{tósz} (dog in Kashubian) is present in the Kashubian-Polish dictionary and its translation is \texttt{pies}. Since the token \texttt{pies} exists in the vocabulary of the source model, we copy the token embedding directly from the source model to the target model.

If the target token is not present in the bilingual dictionary, but is present in the source vocabulary, we also copy its weights directly.

If the target token is not present in either the bilingual dictionary or the source vocabulary, we split the token from the target vocabulary using the source tokenizer. As a result we get the sequence of tokens from the source vocabulary. We obtain the embedding by averaging the token embeddings from the source model.

We use existing bilingual dictionaries for vocabulary matching between the target language and Polish, which we use as the source language. For Silesian, we use the Silling dictionary\footnote{\url{https://silling.org/slownik/}} and for Kashubian, we use the Sloworz dictionary.\footnote{\url{https://sloworz.org/}} Both dictionaries contain about 23 thousand pairs (see Table \ref{tab:data}).

\subsection{Pre-training}
We train the BERT base model using an MLM objective implemented in the HuggingFace library \citep{wolf-etal-2020-transformers}. We either train the model from scratch or fine-tune the HerBERT base model \citep{mroczkowski-etal-2021-herbert} for 150 epochs, with a batch size of 720 examples\footnote{Except for training from scratch, when we use 240 examples. Otherwise the training degenerates.} and a learning rate of $5 \cdot 10^{-4}$. We use a vocabulary of 8,000 tokens for both languages. We leave the rest of the hyperparameters at their default values.

\section{Evaluation}
\label{sec:eval}

\subsection{Masked words prediction}
We evaluate the different variants of pre-training BERT models on the task of predicting masked words. To obtain comparable results between models with different tokenizer vocabulary sizes, we mask and predict the whole word rather than individual tokens. When a word consists of multiple tokens, we predict them causally token by token. We report the accuracy for all words within the validation set of 100 randomly selected Wikipedia articles.

\subsubsection{Results}
We use the multilingual models mBERT \cite{devlin-etal-2019-bert} and XLM-R \cite{conneau-etal-2020-unsupervised} as baselines. XLM-R performs the worst, with 16.15\% for Silesian and 14.95\% for Kashubian. Multilingual BERT performs much better with 32.58\% and 31.10\% respectively. 

Since both Silesian and Kashubian share many similarities with Polish, we use HerBERT, a BERT model trained for the Polish language, as another baseline. Without any fine-tuning on the Silesian/Kashubian corpus, it achieves a low accuracy of 33.03\% for Silesian and 34.69\% for Kashubian (see Table \ref{tab:mlm}). After fine-tuning, the accuracy increases to 59.27\% and 55.88\% respectively.

Next, we train the randomly initialized BERT model with a Silesian or Kashubian tokenizer (and not Polish like in HerBERT). The accuracies of 44.83\% for Silesian and 40.23\% for Kashubian are better than the zero-shot HerBERT baseline, but much lower than the fine-tuned HerBERT.

When the model is initialized with the HerBERT weights, the accuracy increases from 44.83\% to 48.71\% for Silesian and from 40.23\% to 46.08\% for Kashubian. It is still lower than the fine-tuned HerBERT, which illustrates the importance of matching the tokenizer vocabulary between the source and target models and allowing to leverage also the embedding layer of the source pre-trained model.

After matching the vocabularies with bilingual dictionaries (see section \ref{sec:vocab}), the performance of the models increases and surpasses the fine-tuned HerBERT. For Silesian, the accuracy is 1 p.p. higher than with fine-tuning HerBERT (60.27\% vs 59.27\%), and the increase for Kashubian is more than 3 p.p. (59.13\% vs 55.88\%).

\begin{table}[!ht]
\renewcommand*{\arraystretch}{1.3}
\setlength{\tabcolsep}{2.5pt}
\centering
\begin{tabular}{lll|rr}
    \toprule
    \bf{\#} & \bf{Tokenizer} & \bf{Model} & \bf{Silesian} & \bf{Kashubian}\\
    \midrule
    \multicolumn{5}{c}{\bf{Zero-shot}} \\
    \midrule
    1 & mBERT   & mBERT   & 32.58 & 31.10 \\ 
    2 & XLM-R   & XLM-R   & 16.15 & 14.95 \\
    3 & HerBERT & HerBERT & 33.03 & 34.69 \\
    4 & Matched & HerBERT & 25.39 & 28.41 \\
    \midrule
    \multicolumn{5}{c}{\bf{Fine-tuned}} \\
    \midrule
    5 & HerBERT & HerBERT & 59.27 & 55.88 \\
    6 & Not matched & Random & 44.83 & 40.23 \\
    7 & Not matched & HerBERT & 48.71 & 46.08 \\
    8 & Matched & HerBERT & \bf{60.27} & \bf{59.13} \\
    \bottomrule
\end{tabular}
\caption{The accuracy of predicting the masked words for the validation set. Models can use either the \emph{HerBERT} tokenizer or the tokenizer trained on the analyzed language. The latter can be either \emph{matched} using a bilingual dictionary and thus able to use the embedding layer of the HerBERT model, or \emph{not matched} where the embedding layer is randomly initialized. The weights of the BERT model itself can be initialized either \emph{randomly} or using the \emph{HerBERT} model.}
\label{tab:mlm}
\end{table}

\subsection{Passage retrieval}
Additionally, we evaluate the Silesian model on the passage retrieval task. We create a small test set of 50 question-passage pairs from the \emph{Did you know?} section of the Silesian Wikipedia (similar to \citealp{marcinczuk-etal-2013-evaluation}). We use the whole Silesian Wikipedia as a passage corpus (10,605 passages) and evaluate the models using both Accuracy@10 (i.e. there is at least one relevant passage within the top 10 retrieved passages) and NDCG@10 (i.e. the score of each relevant passage within the top 10 retrieved passages depends descendingly on its position, \citealp{ndcg}).

\subsubsection{Evaluated models}
First, we evaluate two existing state-of-the-art models in a zero-shot setup, i.e. without any additional fine-tuning on Silesian data. The Multilingual E5 Base model is a powerful neural retriever trained on both weakly labeled and supervised datasets for over 100 languages (but not Silesian) \citep{wang2022text}. Silver Retriever is the best Polish model for passage retrieval \citep{rybak2023silverretriever}.

Next, we compare two of our models fine-tuned on the Silesian corpus. The standard HerBERT model and the model initialized with HerBERT weights using vocabulary matching (models in rows 5 and 8 of Table \ref{tab:mlm}). Due to the lack of a Silesian training set for this task, we fine-tune the models on the two Polish datasets, PolQA \citep{polqa} and MAUPQA \citep{rybak-2023-maupqa,rybak2023silverretriever}. We use a standard Dense Passage Retriever (DPR, \citealp{karpukhin-etal-2020-dense}) architecture implemented in the Tevatron library \citep{Gao2022TevatronAE}. We fine-tune the models for 1,250 steps, with a batch size of 1,024 and a learning rate of $2 \cdot 10^{-5}$. We leave the rest of the hyperparameters at their default values.

\begin{table}[!ht]
\renewcommand*{\arraystretch}{1.3}
\setlength{\tabcolsep}{4pt}
\centering
\begin{tabular}{ll|rr}
    \toprule
    \bf{\#} & \bf{Model} & \bf{Acc@10} & \bf{NDCG@10}\\
    \midrule
    \multicolumn{4}{c}{\bf{Zero-shot}} \\
    \midrule
    1 & Multilingual E5 Base & 80.00 & 74.79 \\
    2 & Silver Retriever & 84.00 & 73.59 \\
    \midrule
    \multicolumn{4}{c}{\bf{Fine-tuned}} \\
    \midrule
    3 & HerBERT & 92.00 & 79.72 \\
    4 & Matched HerBERT & \bf{96.00} & \bf{81.79} \\
    \bottomrule
\end{tabular}
\caption{The performance of the passage retrieval task for Silesian. \emph{HerBERT} refers to the HerBERT model fine-tuned to the Silesian corpus and \emph{Matched HerBERT} refers to the model that uses vocabulary matching (see Section \ref{sec:vocab}).}
\label{tab:qa}
\end{table}

\subsubsection{Results}
The E5 models score 80\% for accuracy and 74.79\% for NDCG. Silver Retriever achieves comparable results, it scores higher for accuracy (84\%), but lower for NDCG (73.59\%).

Both fine-tuned models outperform the zero-shot retrievers, but the HerBERT model with matched vocabulary achieves much higher results, 96\% vs 92\% for accuracy and 81.79\% vs 79.72\% for NDCG. It proves the effectiveness of the vocabulary-matching technique.

\section{Conclusion}
\label{sec:conclusion}

In this paper, we propose a simple method to extend the capabilities of pre-trained language models, in particular BERT, to low-resource languages through vocabulary matching. Experiments conducted on Silesian and Kashubian languages demonstrate the effectiveness of the proposed method on masked word prediction and passage retrieval tasks.

The contributions of this work include the publication of the first BERT models for the Silesian and Kashubian languages, marking a significant advance in linguistic resources for these languages.

\section{Acknowledgments}
This work was supported by the European Regional Development Fund as a part of 2014–2020 Smart Growth Operational Programme, CLARIN — Common Language Resources and Technology Infrastructure, project no. POIR.04.02.00-00C002/19.

\nocite{*}
\section{Bibliographical References}\label{reference}
\bibliographystyle{lrec_natbib}
\bibliography{lrec-coling2024-example}


\end{document}